%% file: main.tex
\documentclass[10pt,twocolumn,letterpaper]{article}

\usepackage{iccv}
\usepackage{times}
\usepackage{epsfig}
\usepackage{graphicx}
\usepackage{amsmath}
\usepackage{amssymb}
\usepackage{algorithm}
\usepackage{algorithmic}
\usepackage{bbm}
\usepackage{amsfonts}
\usepackage{mathrsfs}
\usepackage{multirow}

\usepackage[pagebackref=true,breaklinks=true,letterpaper=true,colorlinks,bookmarks=false]{hyperref}

\iccvfinalcopy 

\def\iccvPaperID{406} 

\begin{document}

\title{Conservative~ Wasserstein~ Training~ for~ Pose~ Estimation}

\author{Xiaofeng Liu{$^{1,2\dag*}$},~ Yang Zou{$^{1\dag}$},~ Tong Che{$^{3\dag}$},~ Peng Ding{$^{4}$},~ Ping Jia{$^{4}$},~ Jane You{$^{5}$},~ B.V.K. Vijaya Kumar{$^{1}$}\\\vspace{-8pt}{\small~}\\
{$^{1}$}Carnegie Mellon University;~~~ {$^{2}$}Harvard University;~~~ {$^{3}$}MILA \\ {$^{4}$}CIOMP, Chinese Academy of Sciences;~ {$^{5}$}The Hong Kong Polytechnic University\\
{\small{{$^{\dag}$}Contribute equally~~{$^{*}$}Corresponding to: \tt{liuxiaofengcmu@gmail.com}}}
}

\maketitle
\thispagestyle{empty}

\input{1_Abstract.tex}

\input{2_Introduction.tex}

\input{3_RelatedWork.tex}

\input{4_Approach.tex}

\input{4_Approach1.tex}

\input{4_Approach2.tex}

\input{4_Approach3.tex}

\input{5_Experiments.tex}
\input{5_Experiments1.tex}

\input{5_Experiments2.tex}
\input{5_Experiments3.tex}

\input{6_Conclusions.tex}

{\small
\bibliographystyle{ieee_fullname}
\bibliography{egbib}
}

\end{document}

%% file: 1_Abstract.tex
\begin{abstract}
This paper targets the task with discrete and periodic class labels ($e.g.,$ pose/orientation estimation) in the context of deep learning. The commonly used cross-entropy or regression loss is not well matched to this problem as they ignore the periodic nature of the labels and the class similarity, or assume labels are continuous value. We propose to incorporate inter-class correlations in a Wasserstein training framework by pre-defining ($i.e.,$ using arc length of a circle) or adaptively learning the ground metric. We extend the ground metric as a linear, convex or concave increasing function $w.r.t.$ arc length from an optimization perspective. We also propose to construct the conservative target labels which model the inlier and outlier noises using a wrapped unimodal-uniform mixture distribution. Unlike the one-hot setting, the conservative label
makes the computation of Wasserstein distance more challenging. We systematically conclude the practical closed-form solution of Wasserstein distance for pose data with either one-hot or conservative target label. We evaluate our method on head, body, vehicle and 3D object pose benchmarks with exhaustive ablation studies. The Wasserstein loss obtaining superior performance over the current methods, especially using convex mapping function for ground metric, conservative label, and closed-form solution.\end{abstract}

%% file: 2_Introduction.tex
\section{Introduction}

There are some prediction tasks where the output labels are discrete and are periodic. For example, consider the problem of pose estimation. Although pose can be a continuous variable, in practice, it is often discretized $e.g.$, in 5-degree intervals. Because of the periodic nature of pose, a 355-degree label is closer to 0-degree label than the 10-degree label. Thus it is important to consider the periodic and discrete nature of the pose classification problem.

In previous literature, pose estimation is often cast as a multi-class classification problem \cite{raza2018appearance}, a metric regression problem \cite{prokudin2018deep}, or a mixture of both \cite{mahendran2018mixed}.

\begin{figure}[t]
\centering
\includegraphics[width=8.3cm]{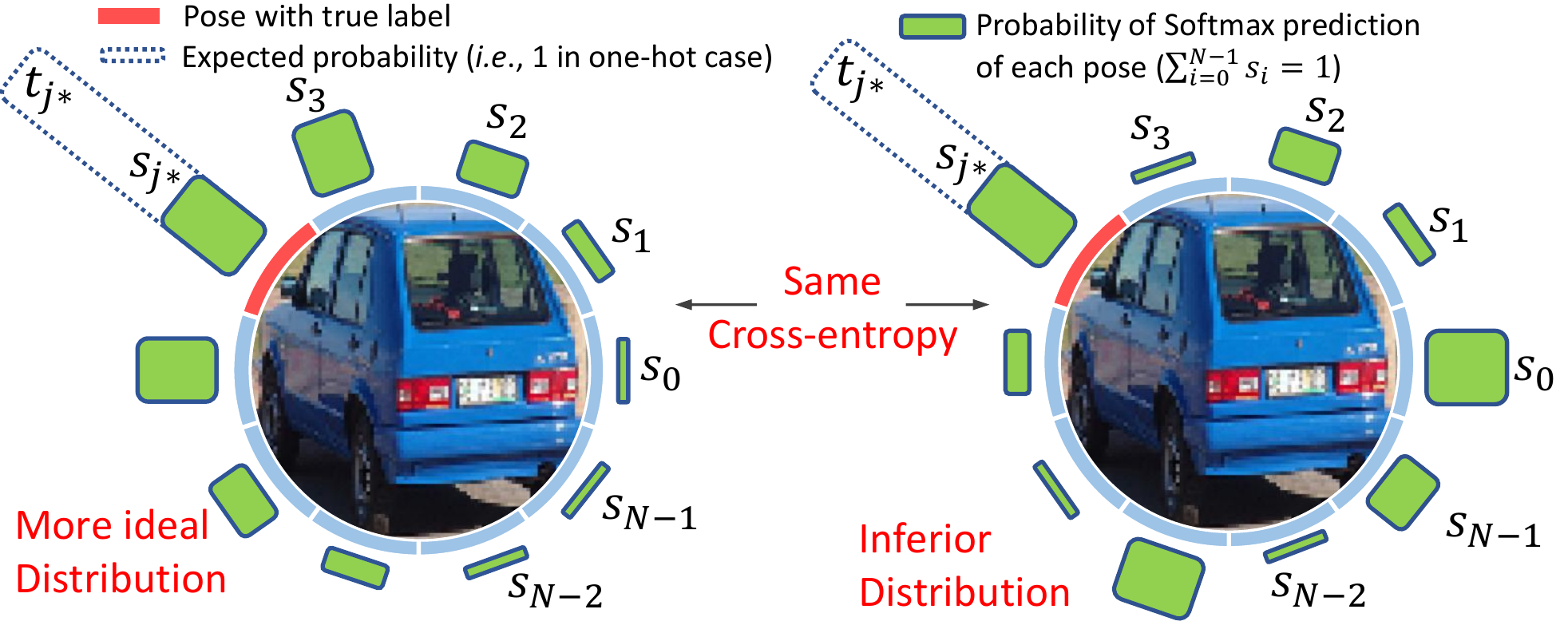}\\
\caption{The limitation of CE loss for pose estimation. The ground truth direction of the car is $t_j \ast$. Two possible softmax predictions (green bar) of the pose estimator have the same probability at $t_j \ast$ position. Therefore, both predicted distributions have the same CE loss. However, the left prediction is preferable to the right, since we desire the predicted probability distribution to be larger and closer to the ground truth class.}\label{fig:1} 
\end{figure}

In a multi-class classification formulation using the cross-entropy (CE) loss, the class labels are assumed to be independent from each other \cite{raza2018appearance,liu2019research,liu2019feature}. Therefore, the inter-class similarity is not properly exploited. For instance, in Fig. \ref{fig:1}, we would prefer the predicted probability distribution to be concentrated near the ground truth class, while the CE loss does not encourage that.

On the other hand, metric regression methods treat the pose as a continuous numerical value \cite{liu2017adaptive,liu2018adaptive,liu2019hard}, although the label of pose itself is discrete. As manifested in \cite{liu2018ordinal,liu2019unimodala,liu2019unimodalb}, learning a regression model using discrete labels will cause over-fitting and exhibit similar or inferior performance compared with classification.

Recent works either use a joint classification and regression loss \cite{mahendran2018mixed} or divide a circle into several sectors with a coarse classification that ignores the periodicity, and then applying regression networks to each sector independently as an ordinal regression problem \cite{hara2017designing}. Unfortunately, none of them fundamentally address the limitations of CE or regression loss in angular data.

In this work, we employ the Wasserstein loss as an alternative for empirical risk minimization. The $1^{st}$ Wasserstein distance is defined as the cost of optimal transport for moving the mass in one distribution to match the target distribution \cite{rubner2000earth,ruschendorf1985wasserstein}. Specifically, we measure the Wasserstein distance between a softmax prediction and its target label, both of which are normalized as histograms. By defining the ground metric as class similarity, we can measure prediction performance in a way that is sensitive to correlations between the classes.

The ground metric can be predefined when the similarity structure is known a priori to incorporate the inter-class correlation, $e.g.,$ the arc length for the pose. We further extend the arc length to its increasing function from an optimization perspective. The exact Wasserstein distance in one-hot target label setting can be formulated as a soft-attention scheme of all prediction probabilities and be rapidly computed. We also propose to learn the optimal ground metric following alternative optimization.  

Another challenge of pose estimation comes from low image quality ($e.g.,$ blurry, low resolution) and the consequent noisy labels. This requires 1) modeling the noise for robust training \cite{liu2019unimodala,liu2019unimodalb} and 2) quantifying the uncertainty of predictions in testing phase \cite{prokudin2018deep}.

Wrongly annotated targets may bias the training process \cite{szegedy2016rethinking,belagiannis2015robust}. We instigate two types of noise. The outlier noise corresponds to one training sample being very distant from others by random error and can be modeled by a uniform distribution \cite{szegedy2016rethinking}. We notice that the pose data is more likely to have inlier noise where the labels are wrongly annotated as the near angles and propose to model it using a unimodal distribution. Our solution is to construct a conservative target distribution by smoothing the one-hot label using a wrapped uniform-unimodal mixture model.

Unlike the one-hot setting, the conservative target distribution makes the computation of Wasserstein distance more advanced because of the numerous possible transportation plans. The $\mathcal{O}(N^3)$ computational complexity for $N$ classes has long been a stumbling block in using Wasserstein distance for large-scale applications. Instead of only obtaining its approximate solution using a $\mathcal{O}(N^2)$ complexity algorithm  \cite{cuturi2013sinkhorn}, 
we systematically analyze the fast closed-form computation of Wasserstein distance for our conservative label when our ground metric is a linear, convex, or concave increasing function $w.r.t.$ the arc length. The linear and convex cases can be solved with linear complexity of $\mathcal{O}(N)$. Our $exact$ solutions are significantly more efficient than the approximate baseline.

The main contributions of this paper are summarized as

$\bullet$ We cast the pose estimation as a Wasserstein training problem. The inter-class relationship of angular data is explicitly incorporated as prior information in our ground metric which can be either pre-defined (a function $w.r.t.$ arc length) or adaptively learned with alternative optimization.

$\bullet$ We model the inlier and outlier error of pose data using a wrapped discrete unimodal-uniform mixture distribution, and regularize the target confidence by transforming one-hot label to conservative target label.

$\bullet$ For either one-hot or conservative target label, we systematically conclude the possible fast closed-form solution when a non-negative linear, convex or concave increasing mapping function is applied in ground metric.  

We empirically validate the effectiveness and generality of the proposed method on multiple challenging benchmarks and achieve the state-of-the-art performance.

%% file: 3_RelatedWork.tex
\section{Related Works}
\noindent\textbf{Pose or viewpoint estimation} has a long history in computer vision \cite{murphy2009head}. It arises in different applications, such as head \cite{murphy2009head}, pedestrian body \cite{raza2018appearance}, vehicle \cite{yang2018hierarchical} and object class \cite{su2015render} orientation/pose estimation. Although these systems are mostly developed independently, they are essentially the same problem in our framework.

The current related literature using deep networks can be divided into two categories. Methods in the first group, such as \cite{rad2017bb8,grabner20183d,zhou2018starmap}, predict keypoints in images and then recover the pose using pre-defined 3D object models. The keypoints can be either semantic \cite{pavlakos20176,wu2016single,massa2016crafting} or the eight corners of a 3D bounding box encapsulating the object \cite{rad2017bb8,grabner20183d}. 

The second category of methods, which are more close to our approach, estimate angular values directly from the image \cite{elhoseiny2016comparative,wang2016viewpoint}. Instead of the typical Euler angle representation for rotations \cite{elhoseiny2016comparative}, biternion representation is chosen in \cite{beyer2015biternion,prokudin2018deep} and inherits the periodicity in its $sin$ and $cos$ operations. However, their setting is compatible with only the regression. Several studies have evaluated the performance of classification and regression-based loss functions and conclude that the classification methods usually outperform the regression ones in pose estimation \cite{massa2016crafting,mahendran2018mixed}. 

These limitations were also found in the recent approaches which combine classification with regression or even triplet loss \cite{mahendran2018mixed,yang2018hierarchical}.

\noindent\textbf{Wasserstein distance} is a measure defined between probability distributions on a given metric space \cite{kolouri2016sliced}. Recently, it attracted much attention in generative models $etc$ \cite{arjovsky2017wasserstein}. \cite{frogner2015learning} introduces it for multi-class multi-label task with a linear model. Because of the significant amount of computing needed to solve the exact distance for general cases, these methods choose the approximate solution, whose complexity is still in $\mathcal{O}(N^2)$ \cite{cuturi2013sinkhorn}. The fast computing of discrete Wasserstein distance is also closely related to SIFT \cite{cha2002measuring} descriptor, hue in HSV or LCH space \cite{cha2002fast} and sequence data \cite{su2017order}. Inspired by the above works, we further adapted this idea to the pose estimation, and encode the geometry of label space by means of the ground matrix. We show that the fast algorithms exist in our pose label structure using the one-hot or conservative target label and the ground metric is not limited to the arc length.

\noindent\textbf{Robust training with noise data} has long been studied for general classification problems \cite{huber2011robust}. Smoothing the one-hot label \cite{szegedy2016rethinking} with a uniform distribution or regularizing the entropy of softmax output \cite{pereyra2017regularizing} are two popular solutions. Some works of regression-based localization model the uncertainty of point position in a plane with a 2D Gaussian distribution \cite{szeto2017click}. \cite{zou2019confidence} propose to regularize self-training with confidence. However, there are few studies for the discrete periodic label. Besides sampling on Gaussian, the Poisson and the Binomial distribution are further discussed to form a unimodal-uniform distribution.

\noindent\textbf{Uncertainty quantification of pose estimation} aims to quantify the reliability of a result $e.g.,$ a confidence distribution of each class rather than a certain angle value for pose data \cite{prokudin2018deep}. A well-calibrated uncertainty is especially important for large systems to assess the consequence of a decision \cite{che2019deep,han2019unsupervised}. \cite{prokudin2018deep} proposes to output numerous sets of the mean and variation of Gaussian/Von-Mises distribution following \cite{beyer2015biternion}. It is unnecessarily complicated and is a somewhat ill-matched formulation as it assumes the pose label is continuous, while it is discrete. We argue that the $softmax$ is a natural function to capture discrete uncertainty, and is compatible with Wasserstein training.

%% file: 4_Approach.tex
\section{Methodology}

We consider learning a pose estimator ${h}_\theta$, parameterized by $\theta$, with $N$-dimensional softmax output unit. It maps a image {\rm\textbf{x}} to a vector ${\rm\textbf{s}}\in\mathbb{R}^N$. We perform learning over a hypothesis space $\mathcal{H}$ of ${h}_\theta$. Given input {\rm\textbf{x}} and its target ground truth one-hot label ${\rm\textbf{t}}$, typically, learning is performed via empirical risk minimization to solve $\mathop{}_{{h}_\theta\in\mathcal{H}}^{\rm min}\mathcal{L}({h}_\theta({\rm\textbf{x}}),{\rm\textbf{t}})$, with a loss $\mathcal{L}(\cdot,\cdot)$ acting as a surrogate of performance measure.

Unfortunately, cross-entropy, information divergence, Hellinger distance and $\mathcal{X}^2$ distance-based loss treat the output dimensions independently \cite{frogner2015learning}, ignoring the similarity structure on pose label space.  

Let ${\rm\textbf{s}}=\left\{s_i\right\}_{i=0}^{N-1}$ be the output of ${h}_\theta({\rm\textbf{x}})$, $i.e.,$ softmax prediction with $N$ classes (angles), and define ${\rm\textbf{t}}=\left\{t_j\right\}_{j=0}^{N-1}$ as the target label distribution, where $i,j\in\left\{0,\cdots,{\small N-1}\right\}$ be the index of dimension (class). Assume class label possesses a ground metric ${\rm\textbf{D}}_{i,j}$, which measures the semantic similarity between $i$-th and $j$-th dimensions of the output. There are $N^2$ possible ${\rm\textbf{D}}_{i,j}$ in a $N$ class dataset and form a ground distance matrix $\textbf{D}\in\mathbb{R}^{N\times N}$. When ${\rm\textbf{s}}$ and ${\rm\textbf{t}}$ are both histograms, the discrete measure of exact Wasserstein loss is defined as \begin{equation}
\mathcal{L}_{\textbf{D}_{i,j}}({\rm{\textbf{s},\textbf{t}}})=\mathop{}_{\textbf{W}}^{{\rm inf}}\sum_{j=0}^{N-1}\sum_{i=0}^{N-1}\textbf{D}_{i,j}\textbf{W}_{i,j} \label{con:df}
\end{equation} where \textbf{W} is the transportation matrix with \textbf{W}$_{i,j}$ indicating the mass moved from the $i^{th}$ point in source distribution to the $j^{th}$ target position. A valid transportation matrix \textbf{W} satisfies: $\textbf{W}_{i,j}\geq 0$; $\sum_{j=0}^{N-1}\textbf{W}_{i,j}\leq s_i$; $\sum_{i=0}^{N-1}\textbf{W}_{i,j}\leq t_j$; $\sum_{j=0}^{N-1}\sum_{i=0}^{N-1}\textbf{W}_{i,j}={\rm min}(\sum_{i=0}^{N-1}s_i,\sum_{j=0}^{N-1}t_j)$.

The ground distance matrix ${\rm\textbf{D}}$ in Wasserstein distance is usually unknown, but it has clear meanings in our application. Its $i,j$-th entry ${\rm\textbf{D}}_{i,j}$ could be the geometrical distance between the $i$-th and $j$-th points in a circle. A possible choice is using the arc length ${d_{i,j}}$ of a circle ($i.e., \ell_1$ distance between the $i$-th and $j$-th points in a circle) as the ground metric $\textbf{D}_{i,j}={d_{i,j}}$.

\begin{equation}
d_{i,j}={\rm min}\left\{|i-j|,N-|i-j|\right\} \label{con:d}
\end{equation}

The Wasserstein distance is identical to the Earth mover's distance when the two distributions have the same total masses ($i.e., \sum_{i=0}^{N-1}s_i=\sum_{j=0}^{N-1}t_j$) and using the symmetric distance $d_{i,j}$ as ${\rm\textbf{D}}_{i,j}$.

This setting is satisfactory for comparing the similarity of SIFT or hue \cite{rubner2000earth}, which do not use a neural network optimization. The previous efficient algorithm usually holds only for $\textbf{D}_{i,j}={d_{i,j}}$. We propose to extend the ground metric in ${\rm\textbf{D}}_{i,j}$ as $f(d_{i,j})$, where $f$ is a positive increasing function $w.r.t.$ $d_{i,j}$.

%% file: 4_Approach1.tex
\subsection{Wasserstein training with one-hot target}

The one-hot encoding is a typical setting for multi-class one-label dataset. The distribution of a target label probability is ${\rm\textbf{t}}=\delta_{j,j^*}$, where $j^*$ is the ground truth class, $\delta_{j,j^*}$ is a Dirac delta, which equals to 1 for $j=j^*$\footnote{\noindent We use $i,j$ interlaced for ${\rm \textbf{s}}$ and ${\rm \textbf{t}}$, since they index the same group of positions in a circle.}, and $0$ otherwise. 

\begin{figure}[t]
\centering
\begin{tabular}{cc}
\includegraphics[height=2.8cm]{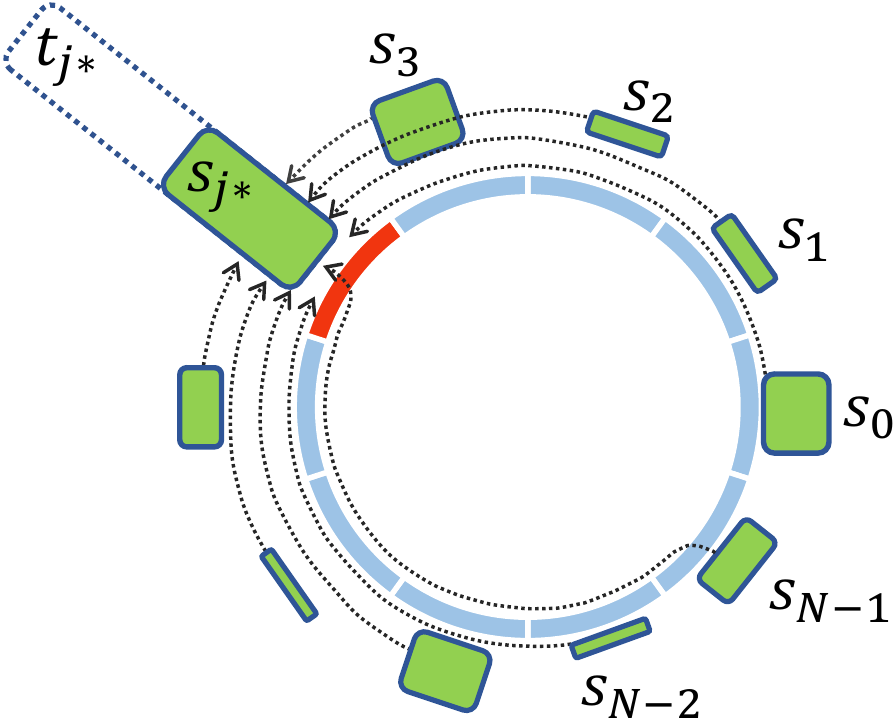}&\includegraphics[height=2.8cm]{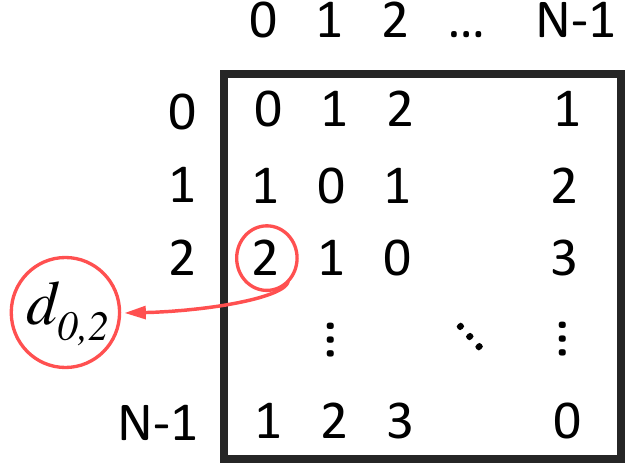}
\end{tabular}
\caption{Left: The only possible transport plan in one-hot target case. Right: the ground matrix using arc length as ground metric.}
\label{fig:2}
\end{figure} 

\noindent\textbf{Theorem 1.} \textit{Assume that} $\sum_{j=0}^{N-1}t_j=\sum_{i=0}^{N-1}s_i$, \textit{and} ${\rm{\textbf{t}}}$ \textit{is a one-hot distribution with} $t_{j^*}=1 ($or $\sum_{i=0}^{N-1}s_i)$\footnote{We note that softmax cannot strictly guarantee the sum of its outputs to be 1 considering the rounding operation. However, the difference of setting $t_{j^*}$ to $1$ or $\sum_{i=0}^{N-1}s_i)$ is not significant in our experiments using the typical format of softmax output which is accurate to 8 decimal places.}, \textit{there is only one feasible optimal transport plan.}

According to the criteria of ${\rm\textbf{W}}$, all masses have to be transferred to the cluster of the ground truth label $j^*$, as illustrated in Fig. \ref{fig:2}. Then, the Wasserstein distance between softmax prediction {\rm{\textbf{s}}} and one-hot target {\rm{\textbf{t}}} degenerates to\begin{equation}
\mathcal{L}_{{\rm\textbf{D}}_{i,j}^{f}}({\rm{\textbf{s},\textbf{t}}})=\sum_{i=0}^{N-1} s_i f(d_{i,j^*}) \label{con:df}
\end{equation} where ${\rm\textbf{D}}_{i,j}^f=f(d_{i,j})$. Practically, $f$ can be an increasing function proper, $e.g., p^{th}$ power of $d_{i,j}$ and Huber function. The exact solution of Eq. \eqref{con:df} can be computed with a complexity of $\mathcal{O}(N)$. The ground metric term $f(d_{i,j^*})$ works as the weights $w.r.t.$ $s_i$, which takes all classes into account following a soft attention scheme \cite{liu2018dependency,liu2019dependency,liu2019permutation}. It explicitly encourages the probabilities distributing on the neighboring classes of $j^*$. Since each $s_i$ is a function of the network parameters, differentiating $\mathcal{L}_{{\rm\textbf{D}}_{i,j}^{f}} w.r.t.$ network parameters yields $\sum_{i=0}^{N-1}s_i'f(d_{i,j^*})$.

In contrast, the cross-entropy loss in one-hot setting can be formulated as $-1{\rm log}s_{j^*}$, which only considers a single class prediction like the hard attention scheme \cite{liu2018dependency,liu2019dependency,liu2019permutation}, that usually loses too much information. Similarly, the regression loss using softmax prediction could be $f(d_{i^*,j^*})$, where $i^*$ is the class with maximum prediction probability.

In addition to the predefined ground metric, we also propose to learn ${\rm\textbf{D}}$ adaptively along with our training following an alternative optimization scheme \cite{liu2018joint}.

\noindent\textbf{Step 1:} Fixing ground matrix ${\rm\textbf{D}}$ to compute $\mathcal{L}_{{\rm\textbf{D}}_{i,j}}({\rm{\textbf{s},\textbf{t}}})$ and updating the network parameters.

\noindent\textbf{Step 2:} Fixing network parameters and postprocessing ${\rm\textbf{D}}$ using the feature-level $\ell_1$ distances between different poses.

We use the normalized second-to-last layer neural response in this round as feature vector, since there is no subsequent non-linearities. Therefore, it is meaningful to average the feature vectors in each pose class to compute their centroid and reconstruct ${{\rm\textbf{D}}_{i,j}}$ using the $\ell_1$ distances between these centroids $\overline{d}_{i,j}$. To avoid the model collapse, we construct the ${{\rm\textbf{D}}_{i,j}=\frac{1}{1+\alpha}\left\{f(\overline{d}_{i,j})+\alpha f(d_{i,j})\right\}}$ in each round, and decrease $\alpha$ from 10 to 0 gradually in the training.

%% file: 4_Approach2.tex
\subsection{Wrapped unimodal-uniform smoothing}

The outlier noise exists in most of data-driven tasks, and can be modeled by a uniform distribution \cite{szegedy2016rethinking}. However, pose labels are more likely to be mislabeled as a close class of the true class. It is more reasonable to construct a unimodal distribution to depict the inlier noise in pose estimation, which has a peak at class $j^*$ while decreasing its value for farther classes. We can sample on a continuous unimodal distribution ($e.g.,$ Gaussian distribution) and followed by normalization, or choose a discrete unimodal distribution ($e.g.,$ Poisson/Binomial distribution).

\noindent\textbf{Gaussian/Von-Mises Distribution} has the probability density function (PDF) $f(x)=\frac{{\rm{exp}}\left\{-(x-\mu)^2/2\sigma^2\right\}}{\sqrt{2\pi\sigma^2}}$ for $x\in [0,\small K]$, where $\mu=K/2$ is the mean, and $\sigma^{2}$ is the variance. Similarly, the Von-Mises distribution is a close approximation to the circular analogue of the normal distribution ($i.e., \small K=N-1$). We note that the geometric loss \cite{su2015render} is a special case, when we set $\xi=1,\eta=0$, $\small{K=N-1}$, remove the normalization and adopt CE loss. Since we are interested in modeling a discrete distribution for target labels, we simply apply a softmax operation over their PDF. Note that the output values are mapped to be defined on the circle.

\noindent\textbf{Poisson Distribution} is used to model the probability of the number of events, $k$ occurring in a particular interval of time. Its probability mass function (PMF) is:\vspace{-5pt}\begin{equation}
p_k=\frac{\lambda^k{\rm{exp}}(-\lambda)}{k!},~~~k= 0, 1, 2, ...,       
\end{equation}\vspace{-2pt}where $\lambda\in \mathbb{R}^+$ is the average frequency of these events. We can sample $K+1$ probabilities ($i.e., 0\leq k\leq K$) on this PMF and followed by normalization for discrete unimodal probability distributions. Since its mean and variation are the same ($i.e., \lambda$), it maybe inflexible to adjust its shape.

\begin{figure}[t]
\centering
\includegraphics[width=8.2cm]{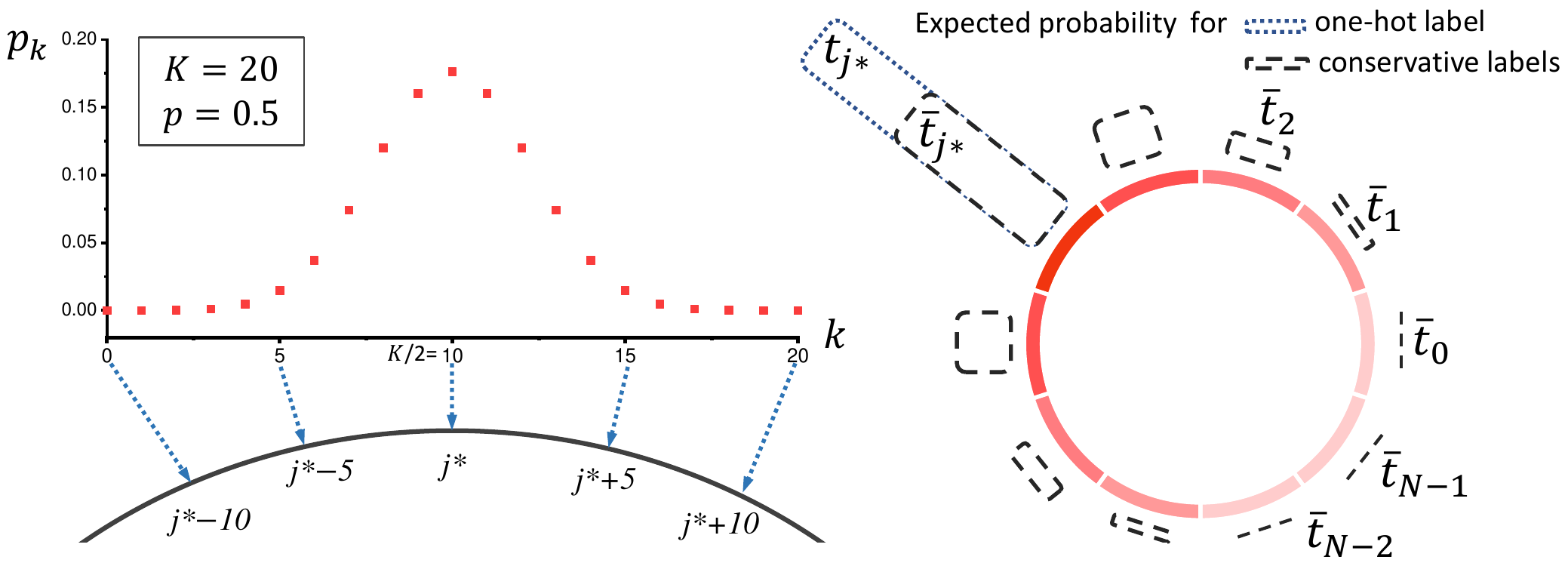}\\
\caption{Left: the wrapping operation with a Binomial distribution ($\small K+1$ is the number of involved classes of unimodal distribution). Right: the distribution of conservative target label.}\label{fig:3}
\end{figure}

\noindent\textbf{Binomial Distribution} is commonly adopted to model the probability of a given number of successes out of a given number of trails $k$ and the success probability $p$.\vspace{-5pt}\begin{equation}
p_k={n\choose k}p^k(1-p)^{n-k},~~~n\in\mathbb{N},~~k=0,1,2, ...,n
\end{equation}\vspace{-2pt}We set $n=K$ to construct a distribution with $K+1$ bins without softmax normalization. Its warp processing with $K=20$ is illustrated in Fig. \ref{fig:3}.

The conservative target distribution ${\rm{{\overline{\textbf{t}}}}}$ is constructed by replacing $t_{j}$ in ${\rm\textbf{t}}$ with $(1-\xi-\eta)t_{j}+\xi p_j +\eta \frac{1}{N}$, which can be regarded as the weighted sum of the original label distribution ${\rm\textbf{t}}$ and a unimodal-uniform mixture distribution. When we only consider the uniform distribution and utilize the CE loss, it is equivalent to label smoothing \cite{szegedy2016rethinking}, a typical mechanism for outlier noisy label training, which encourages the model to accommodate less-confident labels.

By enforcing {\rm\textbf{s}} to form a unimodal-uniform mixture distribution, we also implicitly encourage the probabilities to distribute on the neighbor classes of $j^*$.

%% file: 4_Approach3.tex
\subsection{Wasserstein training with conservative target}

With the conservative target label, the fast computation of Wasserstein distance in Eq. \eqref{con:df} does not apply. A straightforward solution is to regard it as a general case and solve its closed-form result with a complexity higher than $\mathcal{O}(N^3)$ or get an approximate result with a complexity in $\mathcal{O}(N^2)$. The main results of this section are a series of analytic formulation when the ground metric is a nonnegative increasing linear/convex/concave function $w.r.t.$ arc length with a reasonable complexity.\\

\noindent\textbf{3.3.1 Arc length $d_{i,j}$ as the ground metric.}

When we use $d_{i,j}$ as ground metric directly, the Wasserstein loss $\mathcal{L}_{d_{i,j}}(\rm{\textbf{s},\overline{\textbf{t}}})$ can be written as \begin{equation}
\mathcal{L}_{d_{i,j}}{(\rm{{\textbf{s},\overline{\textbf{t}}}})=\mathop{}_{\alpha\in\mathbb{R}}^{inf}}\sum_{j=0}^{N-1}|{\sum_{i=0}^{j}(s_i-\overline{{t}}_i)}-\alpha|\label{con:medi}
\end{equation}

To the best of our knowledge, Eq. \eqref{con:medi} was first developed in \cite{werman1986bipartite}, in which it is proved for sets of points with unitary masses on the circle. A similar conclusion for the Kantorovich-Rubinstein problem was derived in \cite{cabrelli1995kantorovich,cabrelli1998linear}, which is known to be identical to the Wasserstein distance problem when ${{\rm\textbf{D}}_{i,j}}$ is a distance. We note that this is true for $\mathcal{L}_{d_{i,j}}$ (but false for $\mathcal{L}_{{\rm\textbf{D}}^{\rho}}{(\rm{{\textbf{s},\overline{\textbf{t}}}})}$ with $\rho>1$). The optimal $\alpha$ should be the median of the set values $\left\{\sum_{i=0}^{j}(s_i-\overline{{t}}_i), 0\leq j\leq {\scriptsize{N}}-1\right\}$ \cite{pele2008linear}. An equivalent distance is proposed from the circular cumulative distribution perspective \cite{rabin2009statistical}. All of these papers notice that computing Eq. \eqref{con:medi} can be done in linear time ($i.e., \mathcal{O}(N)$) weighted median algorithm (see \cite{villani2003topics} for a review).

We note that the partial derivative of Eq. \eqref{con:medi} $w.r.t.$ $s_n$ is $\sum_{j=0}^{N-1}{\rm{sgn}}(\varphi_j)\sum_{i=0}^{j}(\delta_{i,n}-s_i),$ where $\varphi_j=\sum_{i=0}^{j}(s_i-{\overline{t}_i}),$ and $\delta_{i,n}=1$ when $i=n$. Additional details are given in Appendix B.

~\\
\noindent\textbf{3.3.2 Convex function $ w.r.t.$ $d_{i,j}$ as the ground metric}

Next, we extend the ground metric as an nonnegative increasing and convex function of $d_{i,j}$, and show its analytic formulation. If we compute the probability with a precision $\epsilon$, we will have $M=1/\epsilon$ unitary masses in each distribution. We define the cumulative distribution function of ${\rm{\textbf{s}}}$ and ${\overline{\rm\textbf{t}}}$ and their pseudo-inverses as follows \begin{equation}
\begin{array}{ll}
{\rm{\textbf{S}}}(i)=\sum_{i=0}^{N-1}s_i; ~{\rm{\textbf{S}}}{(m)}^{-1}={\rm{inf}}\left\{i; {\rm{\textbf{S}}}(i)\geq m\right\}\\

{\rm\overline{\textbf{T}}}(i)=\sum_{i=0}^{N-1}\overline{t}_i; ~{\rm\overline{\textbf{T}}}{(m)}^{-1}={\rm{inf}}\left\{i; {\rm\overline{\textbf{T}}}(i)\geq m\right\}\\
             \end{array}
\end{equation} where $m\in \left\{\frac{1}{M},\frac{2}{M},\cdots,1\right\}$. Following the convention ${\rm{\textbf{S}}}(i+N)={\rm{\textbf{S}}}(i)$, ${\rm{\textbf{S}}}$ can be extended to the whole real number, which consider ${\rm{\textbf{S}}}$ as a periodic (or modulo \cite{cha2002measuring}) distribution on $\mathbb{R}$.

\textbf{Theorem 2.} \textit{Assuming the arc length distance $d_{i,j}$ is given by} Eq. \eqref{con:d} \textit{and the ground metric} ${{\rm\textbf{D}}_{i,j}}=f(d_{i,j})$, \textit{with f a nonnegative, increasing and convex function. Then} \begin{equation}
\mathcal{L}_{{\rm\textbf{D}}^{conv}_{i,j}}{(\rm{{\textbf{s},\overline{\textbf{t}}}})=\mathop{}_{\alpha\in\mathbb{R}}^{inf}}\sum_{m=\frac{1}{M}}^{1} f(|{{\rm\small{\textbf{S}}}(m)}^{-1}-{({\rm\small\overline{\textbf{T}}}(m)-\alpha)}^{-1}|)\label{con:conv}  
\end{equation} where $\alpha$ is a to-be-searched transportation constant. A proof of Eq. \eqref{con:conv} $w.r.t.$ the continuous distribution was given in \cite{delon2010fast}, which shows it holds for any couple of desecrate probability distributions. Although that proof involves some complex notions of measure theory, that is not needed in the discrete setting. The proof is based on the idea that the circle can always be ``cut'' somewhere by searching for a $m$, that allowing us to reduce the modulo problem \cite{cha2002measuring} to ordinal case. Therefore, Eq. \eqref{con:conv} is a generalization of the ordinal data. Actually, we can also extend Wasserstein distance for discrete distribution in a line \cite{villani2003topics} as \begin{equation}\sum_{m=\frac{1}{M}}^{1} f(|{{\rm{\textbf{S}}}(m)}^{-1}-{{\rm\overline{\textbf{T}}}(m)}^{-1}|)\label{con:ordinal} \end{equation} where $f$ can be a nonnegative linear/convex/concave increasing function $w.r.t.$ the distance in a line. Eq. \eqref{con:ordinal} can be computed with a complexity of $\mathcal{O}(N)$ for two discrete distributions. When $f$ is a convex function, the optimal $\alpha$ can be found with a complexity of $\mathcal{O}({\rm log}M)$ using the Monge condition\footnote{${\rm\textbf{D}}_{i,j}$+${\rm\textbf{D}}_{i',j'}<{\rm\textbf{D}}_{i,j'}$+${\rm\textbf{D}}_{i',j}$ whenever $i<i'$ and $j<j'$.} (similar to binary search). Therefore, the exact solution of Eq. \eqref{con:conv} can be obtained with $\mathcal{O}(N{\rm log}M)$ complexity. In practice, $\small {\rm log}M$ is a constant (${\rm log}10^8$) according to the precision of softmax predictions, which is much smaller than $N$ (usually $N=360$ for pose data).

Here, we give some measures\footnote{We refer to ``measure'', since a $\rho^{th}$-root normalization is required to get a distance \cite{villani2003topics}, which satisfies three properties: positive definiteness, symmetry and triangle inequality.} using the typical convex ground metric function.

$\mathcal{L}_{{\rm\textbf{D}}_{i,j}^\rho}{(\rm{{\textbf{s},\overline{\textbf{t}}}})}$, the Wasserstein measure using $d^\rho$ as ground metric with $\rho=2,3,\cdots$. The case $\rho=2$ is equivalent to the Cram\'{e}r distance \cite{rizzo2016energy}. Note that the Cram\'{e}r distance is not a distance metric proper. However, its square root is.\begin{equation}
{\rm\textbf{D}}_{i,j}^\rho= d_{i,j}^\rho    
\end{equation}

\vspace{-3pt}
$\mathcal{L}_{{\rm\textbf{D}}_{i,j}^{H\tau}}{(\rm{{\textbf{s},\overline{\textbf{t}}}})}$, the Wasserstein measure using a Huber cost function with a parameter $\tau$.\begin{equation}
{\rm\textbf{D}}_{i,j}^{H\tau}=\left\{
             \begin{array}{ll}
             d_{i,j}^2&{\rm{if}}~d_{i,j}\leq\tau\\
             \tau(2d_{i,j}-\tau)&{\rm{otherwise}}.\\
             \end{array}
             \right.
\end{equation}

~\\
\noindent\textbf{3.3.3 Concave function $w.r.t.$ $d_{i,j}$ as the ground metric}

In practice, it may be useful to choose the ground metric as a nonnegative, concave and increasing function $w.r.t.$ $d_{i,j}$. For instance, we can use the chord length. \begin{equation}
{\rm\textbf{D}}_{i,j}^{chord}=2r~{\rm sin}(d_{i,j}/2r)
\end{equation}where $r=N/2\pi$ is the radius. Therefore, $f(\cdot)$ can be regarded as a concave and increasing function on interval [0,$N$/2] $w.r.t.$ $d_{i,j}$.

It is easy to show that ${\rm\textbf{D}}_{i,j}^{chord}$ is a distance, and thus $\mathcal{L}_{{\rm\textbf{D}}^{chord}}(\rm{\textbf{s},\overline{\textbf{t}}})$ is also a distance between two probability distributions \cite{villani2003topics}. Notice that a property of concave distance is that they do not move the mass shared by the $\rm{\textbf{s}}$ and ${\rm\overline{\textbf{t}}}$ \cite{villani2003topics}. Considering the Monge condition does not apply for concave function, there is no corresponding fast algorithm to compute its closed-form solution. In most cases, we settle for linear programming. However, the simplex or interior point algorithm are known to have at best a $\mathcal{O}(N^{2.5}{\rm{log}}(ND_{max}))$ complexity to compare two histograms on $N$ bins \cite{orlin1993faster,burkard2009society}, where $D_{max}=f(\frac{N}{2})$ is the maximal distance between the two bins.

Although the general computation speed of the concave function is not satisfactory, the step function $f(t)=\mathbbm{1}_{t\neq 0}$ (one every where except at 0) can be a special case, which has significantly less complexity \cite{villani2003topics}. Assuming that the $f(t)=\mathbbm{1}_{t\neq 0}$, the Wasserstein metric between two normalized discrete histograms on $N$ bins is simplified to the $\ell_1$ distance. \begin{equation}
\mathcal{L}_{\mathbbm{1}{d_{i,j}\neq 0}}{(\rm{{\textbf{s},\overline{\textbf{t}}}})}=\frac{1}{2}\sum_{i=0}^{N-1}{|{\rm{s}}_i-{\rm{\overline{t}}}_i|}=\frac{1}{2}||{\rm{\textbf{s}}}-{\rm{\overline{\textbf{t}}}}||_1
\end{equation}where $||\cdot||_1$ is the discrete $\ell_1$ norm.  

Unfortunately, its fast computation is at the cost of losing the ability to discriminate the difference of probability in a different position of bins.

%% file: 5_Experiments.tex
\section{Experiments}

\begin{table}[t]  
\scriptsize
\renewcommand\arraystretch{1.2}
\label{tab:different_nets}
\begin{center}
\begin{tabular}{|c|c|c|c|c|}
\cline{1-2}\cline{4-5}

{Method}&MAAD&\scriptsize{~}&{Method}&MAAD\\\cline{1-2}\cline{4-5}

BIT\cite{beyer2015biternion}&25.2$^{\circ}$&\scriptsize{~}&A-${\mathcal{L}_{d_{i,j}}}{(\rm{{\textbf{s},{\textbf{t}}}})}$&17.5$^{\circ}$\\\cline{1-2}\cline{4-5}

DDS\cite{prokudin2018deep}$^\text{\dag}$&23.7$^{\circ}$&\scriptsize{~}&A-${\mathcal{L}_{{\rm\textbf{D}}_{i,j}^2}}{(\rm{{\textbf{s},{\textbf{t}}}})}$&\underline{17.3}$^{\circ}$\\\cline{1-2}\cline{4-5}

$\mathcal{L}_{d_{i,j}}(\rm{\textbf{s},{\textbf{t}}})$&18.8$^{\circ}$&\scriptsize{~}& $\approx\mathcal{L}_{d_{i,j}}(\rm{\textbf{s},{\textbf{t}}})$&19.0$^{\circ}$\\\cline{1-2}\cline{4-5}

$\mathcal{L}_{{\rm\textbf{D}}_{i,j}^2}{(\rm{{\textbf{s},{\textbf{t}}}})}$&\textbf{17.1}$^{\circ}$&\scriptsize{~}&$\approx\mathcal{L}_{{\rm\textbf{D}}_{i,j}^2}{(\rm{{\textbf{s},{\textbf{t}}}})}$&17.8$^{\circ}$\\\cline{1-2}\cline{4-5}

$\mathcal{L}_{{\rm\textbf{D}}_{i,j}^{chord}}{(\rm{{\textbf{s},{\textbf{t}}}})}$&$19.1^{\circ}$&\scriptsize{~}&$\approx\mathcal{L}_{{\rm\textbf{D}}_{i,j}^{chord}}{(\rm{{\textbf{s},{\textbf{t}}}})}$&19.5$^{\circ}$\\\cline{1-2}\cline{4-5}

\end{tabular}\label{con:1}
\end{center}
\caption{Results on CAVIAR head pose dataset (the lower MAAD the better).$^\text{\dag}$ Our implementation based on their publicly available codes. The best are in bold while the second best are underlined.}
\end{table}

In this section, we show the implementation details and experimental results on the head, pedestrian body, vehicle and 3D object pose/orientation estimation tasks. To illustrate the effectiveness of each setting choice and their combinations, we give a series of elaborate ablation studies along with the standard measures.

We use the prefix A and $\approx$ denote the adaptively ground metric learning (in Sec. 3.1) and approximate computation of Wasserstein distance \cite{cuturi2013sinkhorn,frogner2015learning} respectively. ${(\rm{{\textbf{s},{\textbf{t}}}})}$ and ${(\rm{{\textbf{s},\overline{\textbf{t}}}})}$ refer to using one-hot or conservative target label. For instance, $\mathcal{L}_{d_{i,j}}(\rm{\textbf{s},{\textbf{t}}})$ means choosing Wasserstein loss with arc length as ground metric and using one-hot target label.

%% file: 5_Experiments1.tex
\subsection{Head pose}
Following \cite{beyer2015biternion,prokudin2018deep}, we choose the occluded version of CAVIAR dataset \cite{fisher2005caviar} and construct the training, validation and testing using 10802, 5444 and 5445 identity-independent images respectively. Since the orientation of gaze is coarsely labeled, and almost 40\% training samples lie within ${\pm} 4^{\circ}$ of the four canonical orientations, regression-based methods \cite{beyer2015biternion,prokudin2018deep} are inefficient.

For fair comparison, we use the same deep batch normalized VGG-style \cite{simonyan2014very} backbone as in \cite{beyer2015biternion,prokudin2018deep}. Instead of a sigmoid unit in their regression model, the last layer is set to a softmax layer with 8 ways for Right, Right-Back, Back, Left-Back, Left, Left-Front, Front and Right-Front poses. 

The metric used here is the mean absolute angular deviation (MAAD), which is widely adopted for angular regression tasks. The results are summarized in Table \textcolor{red}{1}. The Wasserstein training boosts the performance significantly. Using convex $f$ can further improve the result, while the losses with concave $f$ are usually inferior to the vanilla Wasserstein loss with arc length as the ground metric. The adaptive ground metric learning is helpful for the $\mathcal{L}_{d_{i,j}}(\rm{\textbf{s},{\textbf{t}}})$, but not necessary when we extend the ground metric to the square of $d_{i,j}$.

We also note that the exact Wasserstein distances are consistently better than their approximate counterparts \cite{cuturi2013sinkhorn}. More appealingly, in the training stage, $\mathcal{L}_{d_{i,j}}(\rm{\textbf{s},{\textbf{t}}})$ is 5$\times$ faster than $\approx\mathcal{L}_{d_{i,j}}(\rm{\textbf{s},{\textbf{t}}})$ and 3$\times$ faster than conventional regression-based method \cite{prokudin2018deep} to achieve the convergence.

%% file: 5_Experiments2.tex
\subsection{Pedestrians orientation}

\begin{figure}[t]
\centering
\includegraphics[width=8.4cm]{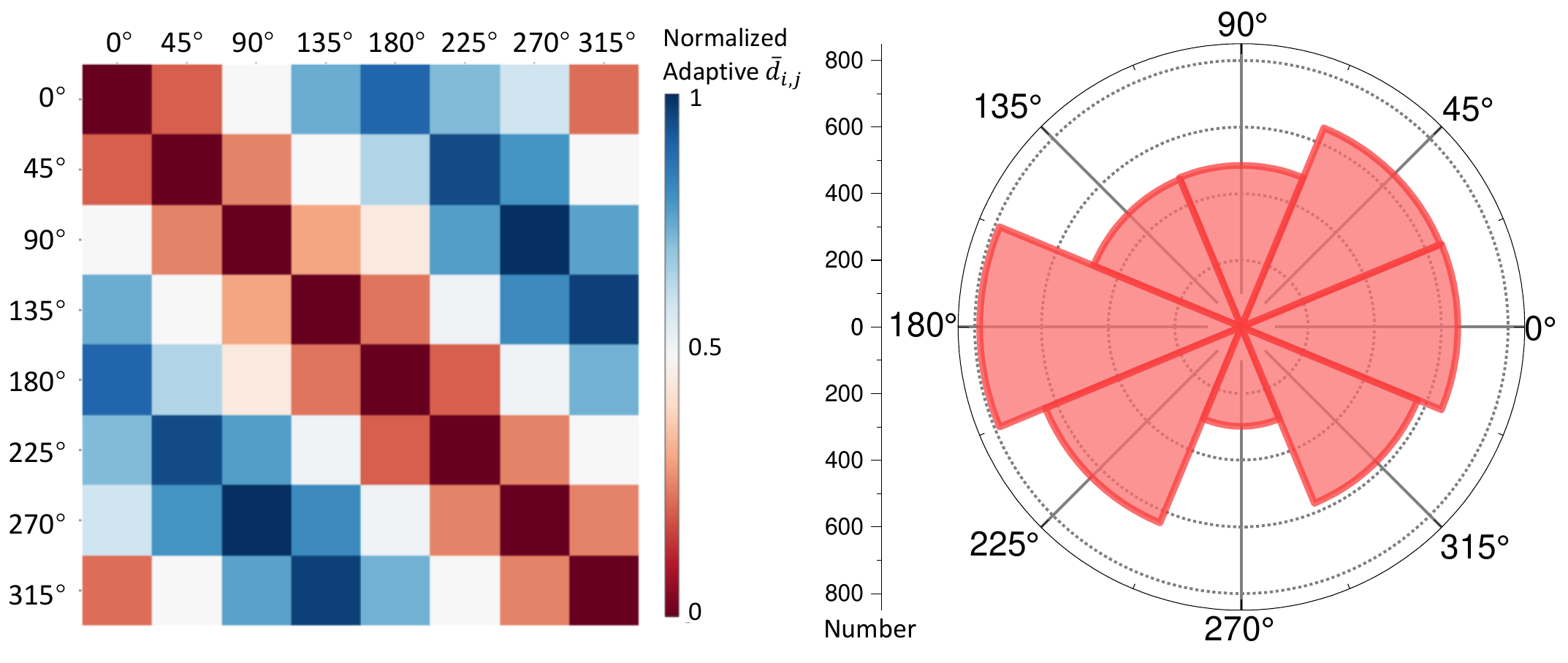}\\
\caption{Normalized adaptively learned ground matrix and polar histogram $w.r.t.$ the number of training samples in TUD dataset.}\label{fig:4}
\end{figure}

The TUD multi-view pedestrians dataset \cite{andriluka2010monocular} consists of 5,228 images along with bounding boxes. Its original annotations are relatively coarse, with only eight classes. We adopt the network in \cite{raza2018appearance} and show the results in Table \textcolor{red}{2}. Our methods, especially the $\mathcal{L}_{{\rm\textbf{D}}_{i,j}^2}{(\rm{{\textbf{s},\overline{\textbf{t}}}})}$ outperform the cross-entropy loss-based approach in all of the eight classes by a large margin. The improvements in the case of binomial-uniform regularization ($\xi=0.1,\eta=0.05,K=4,p=0.5$) seems limited for 8 class setting, because each pose label covers 45$^\circ$ resulting in relatively low noise level.

\begin{table}[t] 
\renewcommand\arraystretch{1.2}
\scriptsize
\label{tab:different_nets}
\begin{center}
\begin{tabular}{|c|c|c|c|c|c|c|c|c|}
\hline
Method&0$^{\circ}$&$45^{\circ}$&$90^{\circ}$&135$^{\circ}$&180$^{\circ}$&225$^{\circ}$&270$^{\circ}$&315$^{\circ}$\\\hline\hline

CE loss\cite{raza2018appearance}&0.90&0.96&0.92&\textbf{1.00}&0.92&0.88&0.89&0.95\\\hline\hline

$\mathcal{L}_{d_{i,j}}(\rm{\textbf{s},{\textbf{t}}})$&0.93&0.97&0.95&\textbf{1.00}&\textbf{0.96}&0.91&0.91&0.95\\\hline

A-${\mathcal{L}_{d_{i,j}}}{(\rm{{\textbf{s},{\textbf{t}}}})}$&0.94&0.97&\textbf{0.96}&\textbf{1.00}&0.95&0.92&0.91&\textbf{0.96}\\\hline

$\mathcal{L}_{{\rm\textbf{D}}_{i,j}^2}{(\rm{{\textbf{s},{\textbf{t}}}})}$&\textbf{0.95}&0.97&\textbf{0.96}&\textbf{1.00}&\textbf{0.96}&0.92&0.91&\textbf{0.96}\\\hline 

CE loss${(\rm{{\textbf{s},\overline{\textbf{t}}}})}$&{0.90}&{0.96}&{0.94}&\textbf{1.00}&{0.92}&{0.90}&{0.90}&{0.95}\\\hline

$\mathcal{L}_{{\rm\textbf{D}}_{i,j}^2}{(\rm{{\textbf{s},\overline{\textbf{t}}}})}$&\textbf{0.95}&\textbf{0.98}&\textbf{0.96}&\textbf{1.00}&\textbf{0.96}&\textbf{0.93}&\textbf{0.92}&\textbf{0.96}\\\hline

\end{tabular}\label{tab:2}
\end{center}
\caption{Class-wise accuracy for TUD pedestrian orientation estimation with 8 pose setting (the higher the better).}
\end{table}

\begin{table}[t]  
\scriptsize
\renewcommand\arraystretch{1.2}
\label{tab:different_nets}
\begin{center}
\begin{tabular}{|c|c|c|c|}
\hline
Method&Mean AE&{$Acc_{\frac{\pi}{8}}$}&{$Acc_{\frac{\pi}{4}}$}\\\hline\hline
RTF\cite{hara2017growing}&34.7&0.686&0.780\\\hline    
SHIFT\cite{hara2017designing}&22.6&0.706&0.861\\\hline\hline  

${\mathcal{L}_{d_{i,j}}}{(\rm{{\textbf{s},{\textbf{t}}}})}$&19.1&0.748&0.900\\\hline
A-${\mathcal{L}_{d_{i,j}}}{(\rm{{\textbf{s},{\textbf{t}}}})}$&20.5&0.723&0.874\\\hline

$\mathcal{L}_{{\rm\textbf{D}}_{i,j}^2}{(\rm{{\textbf{s},{\textbf{t}}}})}$&18.5&0.756&0.905\\\hline

$\mathcal{L}_{{\rm\textbf{D}}_{i,j}^2}~|$ SHIFT ${(\rm{{\textbf{s},\overline{\textbf{t}}}})}_G$&\underline{16.4} $|$ 20.1&\underline{0.764} $|$ 0.724&\underline{0.909} $|$ 0.874\\\hline    

$\mathcal{L}_{{\rm\textbf{D}}_{i,j}^2}~|$ SHIFT ${(\rm{{\textbf{s},\overline{\textbf{t}}}})}_P$&17.7 $|$ 20.8&0.760 $|$ 0.720&0.907 $|$ 0.871\\\hline   

$\mathcal{L}_{{\rm\textbf{D}}_{i,j}^2}~|$ SHIFT ${(\rm{{\textbf{s},\overline{\textbf{t}}}})}_B$&\textbf{16.3} $|$ 20.1&\textbf{0.766} $|$ 0.723&\textbf{0.910} $|$ 0.875\\\hline\hline    

Human \cite{hara2017designing}&9.1&0.907&0.993\\\hline    
    
\end{tabular}\label{tab:3}
\end{center}
\caption{Results on TUD pedestrian orientation estimation $w.r.t.$ Mean Absolute Error in degree (the lower the better) and {$Acc_{\frac{\pi}{8}}$},{$Acc_{\frac{\pi}{4}}$} (the higher the better). $|$ means ``or''. The suffix $G,P,B$ refer to Gaussian, Poison and Binomial-uniform mixture conservative target label, respectively.}
\end{table}

The adaptive ground metric learning can contribute to higher accuracy than the plain $\mathcal{L}_{d_{i,j}}(\rm{\textbf{s},{\textbf{t}}})$. Fig. \ref{fig:4} provides a visualization of the adaptively learned ground matrix. The learned $\overline{d}_{i,j}$ is slightly larger than ${d}_{i,j}$ when limited training samples are available in the related classes, $e.g., {d}_{225^{\circ},180^{\circ}}<{d}_{225^{\circ},270^{\circ}}$. A larger ground metric value may emphasize the class with fewer samples in the training.  

We also utilize the 36-pose labels provided in \cite{hara2017growing,hara2017designing}, and adapt the backbone from \cite{hara2017designing}. We report the results $w.r.t.$ mean absolute error and accuracy at $\frac{\pi}{8}$ and $\frac{\pi}{4}$ in Table \textcolor{red}{3}, which are the percentage of images whose pose error is less than $\frac{\pi}{8}$ and $\frac{\pi}{4}$, respectively. Even the plain $\mathcal{L}_{d_{i,j}}(\rm{\textbf{s},{\textbf{t}}})$ outperforms SHIFT \cite{hara2017designing} by 4.4\% and 3.9\% $w.r.t.$ $Acc\frac{\pi}{8}$ and $Acc\frac{\pi}{4}$. Unfortunately, the adaptive ground metric learning is not stable when we scale the number of class to 36. 

The disagreement of human labeling is significant in 36 class setting. In such a case, our conservative target label is potentially helpful. The discretized Gaussian distribution ($\xi=0.1,\eta=0.05,\mu=5,\sigma^2=2.5$) and Binomial distribution ($\xi=0.1,\eta=0.05,K=10,p=0.5$) show similar performance, while the Poisson distribution ($\xi=0.1,\eta=0.05,K=10,\lambda=5$) appears less competitive. Note that the variance of Poisson distribution is equal to its mean $\lambda$, and it approximates a symmetric distribution with a large $\lambda$. Therefore, it is not easy to control the shape of target distribution. Our $\small \mathcal{L}_{{\rm\textbf{D}}_{i,j}^2}{(\rm{{\textbf{s},\overline{\textbf{t}}}})}_B$ outperforms \cite{hara2017designing} by 6.3$^\circ$, 6\% and 4.9\% in terms of Mean AE, $Acc\frac{\pi}{8}$ and $Acc\frac{\pi}{4}$.

\subsection{Vehicle orientation}

The EPFL dataset \cite{ozuysal2009pose} contains 20 image sequences of 20 car types at a show. We follow \cite{hara2017designing} to choose ResNet-101 \cite{he2016deep} as the backbone and use 10 sequences for training and the other 10 sequences for testing. As shown in Table \textcolor{red}{4}, the Huber function ($\tau=10$) can be beneficial for noisy data learning, but the improvements appear to be not significant after we have modeled the noise in our conservative target label with Binomial distribution ($\xi=0.2,\eta=0.05,K=30,p=0.5$). Therefore, we would recommend choosing $\mathcal{L}_{{\rm\textbf{D}}_{i,j}^2}$ and Binomial-uniform mixture distribution as a simple yet efficient combination. The model is not sensitive to the possible inequality of $\sum_{i=0}^{N-1}t_i$ and $\sum_{i=0}^{N-1}s_i$ caused by numerical precision.

\begin{figure}[t]
\centering
\includegraphics[width=8.3cm]{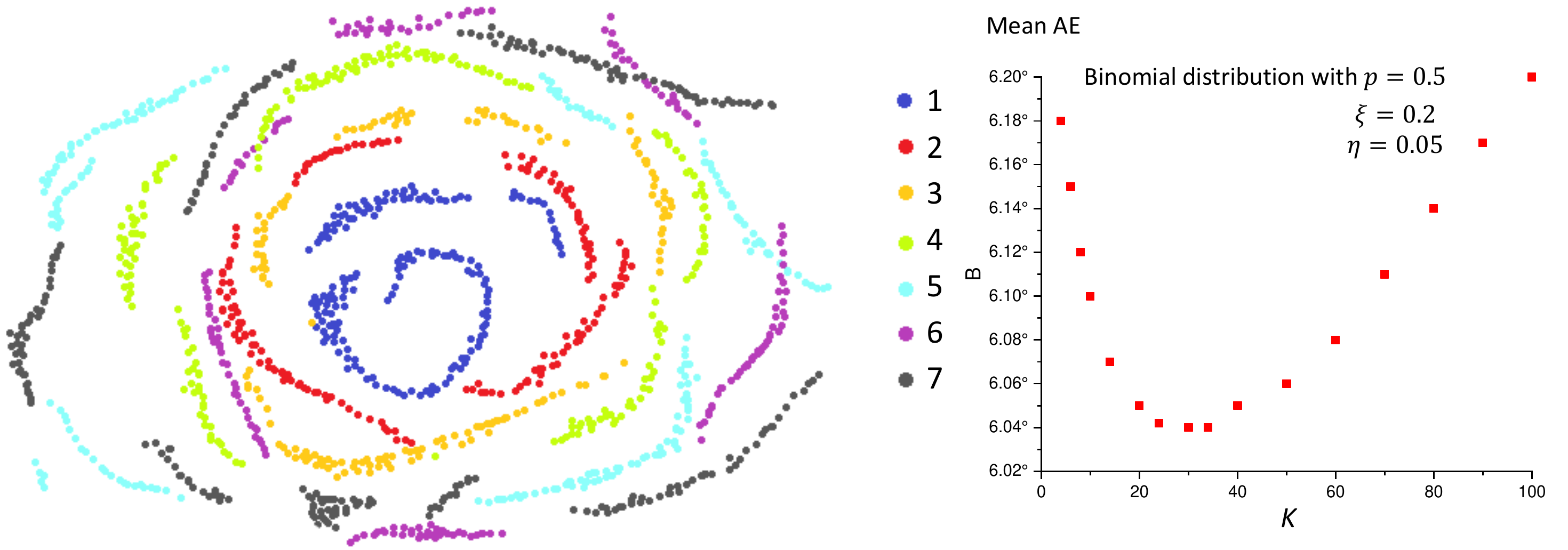}\\
\caption{Left: The second-to-last layer feature of the 7 sequences in EPFL testing set with t-SNE mapping (not space position/angle). Right: Mean AE as a function of $K$ for the Binomial distribution showing that the hyper-parameter $K$ matters.}\label{fig:5}
\end{figure}

Besides, we visualize the second-to-last layer representation of some sequences in Fig. \ref{fig:5} left. As shown in Fig. \ref{fig:5} right, the shape of Binomial distribution is important for performance. It degrades to one-hot or uniform distribution when $K=0$ or a large value. All of the hyper-parameters in our experiments are chosen via grid searching. We see a 27.8\% Mean AE decrease from \cite{hara2017designing} to $\small \mathcal{L}_{{\rm\textbf{D}}_{i,j}^2}{(\rm{{\textbf{s},\overline{\textbf{t}}}})}$, and 33\% for Median AE.

\begin{table}  
\scriptsize
\renewcommand\arraystretch{1.2}
\label{tab:different_nets}
\begin{center}
\begin{tabular}{|c|c|c|}
\hline
Method&Mean AE&Median AE\\\hline\hline
  
HSSR\cite{yang2018hierarchical}&20.30&3.36\\\hline     
SMMR\cite{huang2017soft}&12.61&3.52\\\hline
SHIFT\cite{hara2017designing}&9.86&3.14\\\hline\hline
$\mathcal{L}_{d_{i,j}}{(\rm{{\textbf{s},{\textbf{t}}}})}|{(\rm{{\textbf{s},\overline{\textbf{t}}}})}$&6.46 $|$ 6.30&2.29 $|$ 2.18\\\hline  

$\mathcal{L}_{d_{i,j}}{(\rm{{\textbf{s},{\textbf{t}}}})}|{(\rm{{\textbf{s},\overline{\textbf{t}}}})},t_j*=\sum_{i=0}^{N-1}s_i$$^\text{\dag}$&6.46 $|$ 6.30&2.29 $|$ 2.18\\\hline  

$\mathcal{L}_{{\rm\textbf{D}}_{i,j}^2}{(\rm{{\textbf{s},{\textbf{t}}}})}|{(\rm{{\textbf{s},\overline{\textbf{t}}}})}$&6.23 $|$ \textbf{6.04}&2.15 $|$ \underline{2.11}\\\hline  

$\mathcal{L}_{{\rm\textbf{D}}_{i,j}^2}{(\rm{{\textbf{s},{\textbf{t}}}})}|{(\rm{{\textbf{s},\overline{\textbf{t}}}})},t_j*=\sum_{i=0}^{N-1}s_i$$^\text{\dag}$&6.23 $|$ \textbf{6.04}&2.15 $|$ \underline{2.11}\\\hline  

$\mathcal{L}_{{\rm\textbf{D}}_{i,j}^3}{(\rm{{\textbf{s},{\textbf{t}}}})}|{(\rm{{\textbf{s},\overline{\textbf{t}}}})}$&6.47 $|$ 6.29&2.28 $|$ 2.20\\\hline

$\mathcal{L}_{{\rm\textbf{D}}_{i,j}^{H\tau}}{(\rm{{\textbf{s},{\textbf{t}}}})}|{(\rm{{\textbf{s},\overline{\textbf{t}}}})}$&6.20 $|$ \textbf{6.04}&2.14 $|$ \textbf{2.10}\\\hline    

\end{tabular}\label{tab:4}
\end{center}
\caption{Results on EPFL $w.r.t.$ Mean and Median Absolute Error in degree (the lower the better). $|$ means ``or''.$^\text{\dag}$ denotes we assign $t_j*=\sum_{i=0}^{N-1}s_i$, and $t_j*=1$ in all of the other cases.}
\end{table}

\begin{table*}[t]  
\tiny
\label{tab:different_nets}
\begin{center}
\begin{tabular}{|c|c|c|cccccccccccc|c|}
\cline{3-16}
\multicolumn{2}{c|}{~}&backbone&aero&bike&boat&bottle&bus&car&chair&table&mbike&sofa&train&tv&mean\\\hline

\multirow{12}*{\rotatebox{90}{$Acc_{\frac{\pi}{6}}$}}&Tulsiani $et~al.$ \cite{tulsiani2015viewpoints}&AlexNet+VGG&0.81&0.77&0.59&0.93&\textbf{0.98}&0.89&0.80&0.62&{0.88}&0.82&0.80&0.80&0.8075\\
    &Su $et~al.$ \cite{su2015render}&AlexNet$^\text{\dag}$&0.74&0.83&0.52&0.91&0.91&0.88&0.86&0.73&0.78&0.90&0.86&\textbf{0.92}&0.82\\
    &Mousavian $et~al.$ \cite{mousavian20173d}&VGG&0.78&0.83&0.57&0.93&0.94&0.90&0.80&0.68&0.86&0.82&0.82&0.85&0.8103\\   
  & Pavlakos $et~al.$ \cite{pavlakos20176}&Hourglass&0.81&0.78&0.44&0.79&0.96&0.90&0.80&N/A&N/A&0.74&0.79&0.66&N/A\\ 
   
   &Mahendran $et~al.$ \cite{mahendran2018mixed}&ResNet50$^\text{\dag}$&0.87&0.81&0.64&\textbf{0.96}&\underline{0.97}&\underline{0.95}&\textbf{0.92}&0.67&0.85&\textbf{0.97}&0.82&0.88&0.8588\\
            &Grabner $et~.al.$ \cite{grabner20183d}&ResNet50&0.83&0.82&0.64&0.95&\underline{0.97}&0.94&0.80&0.71&0.88&0.87&0.80&0.86&0.8392\\
            
    &Zhou $et~al.$ \cite{zhou2018starmap}&ResNet18&0.82&\textbf{0.86}&0.50&0.92&\underline{0.97}&0.92&0.79&0.62&0.88&0.92&0.77&0.83&0.8225\\
    &Prokudin $et~al.$ \cite{prokudin2018deep}&InceptionResNet&0.89&0.83&0.46&\textbf{0.96}&0.93&0.90&0.80&0.76&0.90&\textbf{0.90}&0.82&\underline{0.91}&0.84\\\cline{2-16}

&${\mathcal{L}_{d_{i,j}}}{(\rm{{\textbf{s},{\textbf{t}}}})}$&ResNet50&0.88&0.79&\underline{0.67}&0.93&0.96&\textbf{0.96}&0.86&0.73&0.86&0.91&\textbf{0.89}&0.87&0.8735\\

&$\mathcal{L}_{{\rm\textbf{D}}_{i,j}^2}{(\rm{{\textbf{s},{\textbf{t}}}})}$&ResNet50&0.89&\underline{0.84}&\underline{0.67}&\textbf{0.96}&0.95&\underline{0.95}&0.87&0.75&0.88&0.92&\underline{0.88}&0.89&0.8832\\

&$\mathcal{L}_{{\rm\textbf{D}}_{i,j}^{H\tau}}{(\rm{{\textbf{s},{\textbf{t}}}})}$&ResNet50&\underline{0.90}&0.82&\textbf{0.68}&0.95&\underline{0.97}&0.94&\underline{0.89}&\underline{0.76}&0.88&\underline{0.93}&0.87&0.88&\underline{0.8849}\\

&$\mathcal{L}_{{\rm\textbf{D}}_{i,j}^2}{(\rm{{\textbf{s},\overline{\textbf{t}}}})}$&ResNet50&\textbf{0.91}&0.82&\underline{0.67}&\textbf{0.96}&\underline{0.97}&\underline{0.95}&\underline{0.89}&\textbf{0.79}&\textbf{0.90}&\underline{0.93}&0.85&0.90&\textbf{0.8925}\\\hline\hline

\multirow{7}*{\rotatebox{90}{$Acc_{\frac{\pi}{18}}$}}&Zhou $et~al.$ \cite{zhou2018starmap}&ResNet18&0.49&0.34&0.14&0.56&\textbf{0.89}&\textbf{0.68}&0.45&0.29&0.28&\textbf{0.46}&0.58&0.37&0.4818\\\cline{2-16}
     
&${\mathcal{L}_{d_{i,j}}}{(\rm{{\textbf{s},{\textbf{t}}}})}$&ResNet50&0.48&0.64&\textbf{0.20}&\textbf{0.60}&0.83&0.62&0.42&0.37&0.32&0.42&0.58&0.39&0.5020\\

&${\mathcal{L}_{d_{i,j}}}{(\rm{{\textbf{s},\overline{\textbf{t}}}})}$&ResNet50&0.48&0.65&\underline{0.19}&0.58&0.86&0.64&0.45&0.38&\underline{0.35}&0.41&0.55&0.36&0.5052\\

&$\mathcal{L}_{{\rm\textbf{D}}_{i,j}^2}{(\rm{{\textbf{s},{\textbf{t}}}})}$&ResNet50&0.49&0.63&0.18&0.56&0.85&\underline{0.67}&\underline{0.47}&\textbf{0.41}&0.26&0.43&\textbf{0.62}&0.38&0.5086\\

&$\mathcal{L}_{{\rm\textbf{D}}_{i,j}^2}{(\rm{{\textbf{s},\overline{\textbf{t}}}})}$&ResNet50&\underline{0.51}&0.65&\underline{0.19}&\underline{0.59}&0.86&0.63&\textbf{0.48}&\underline{0.40}&0.28&0.41&0.57&\underline{0.40}&\underline{0.5126}\\

&$\mathcal{L}_{{\rm\textbf{D}}_{i,j}^{H\tau}}{(\rm{{\textbf{s},{\textbf{t}}}})}$&ResNet50&\textbf{0.52}&\textbf{0.67}&0.16&0.58&\underline{0.88}&\underline{0.67}&0.45&0.33&0.25&0.44&\underline{0.61}&0.35&0.5108\\

&$\mathcal{L}_{{\rm\textbf{D}}_{i,j}^{H\tau}}{(\rm{{\textbf{s},\overline{\textbf{t}}}})}$&ResNet50&0.50&\underline{0.66}&0.17&0.55&0.85&0.65&0.46&\underline{0.40}&\textbf{0.38}&\underline{0.45}&0.59&\textbf{0.41}&\textbf{0.5165}\\\hline\hline

\multirow{12}*{\rotatebox{90}{$MedErr$}}&Tulsiani $et~al.$ \cite{tulsiani2015viewpoints}&AlexNet+VGG&13.8&17.7&21.3&12.9&5.8&9.1&14.8&15.2&14.7&13.7&8.7&15.4&13.59\\

    &Su $et~al.$ \cite{su2015render}&AlexNet&15.4&14.8&25.6&9.3&3.6&6.0&9.7&10.8&16.7&9.5&6.1&12.6&11.7\\
    &Mousavian $et~al.$ \cite{mousavian20173d}&VGG&13.6&12.5&22.8&8.3&3.1&5.8&11.9&12.5&12.3&12.8&6.3&11.9&11.15\\
    &Pavlakos $et~al.$ \cite{pavlakos20176}&Hourglass&11.2&15.2&37.9&13.1&4.7&6.9&12.7&N/A&N/A&21.7&9.1&38.5&N/A\\
    &Mahendran $et~al.$ \cite{mahendran2018mixed}&ResNet50&\textbf{8.5}&14.8&20.5&7.0&3.1&5.1&\textbf{9.3}&11.3&14.2&10.2&5.6&11.7&11.10\\
    &Grabner $et~al.$ \cite{grabner20183d}&ResNet50&10.0&15.6&\textbf{19.1}&8.6&3.3&5.1&13.7&11.8&12.2&13.5&6.7&11.0&10.88\\
    &Zhou $et~al.$ \cite{zhou2018starmap}&ResNet18&10.1&14.5&30.3&9.1&3.1&6.5&11.0&23.7&14.1&11.1&7.4&13.0&10.4\\ &Prokudin $et~al.$ \cite{prokudin2018deep}&InceptionResNet&\underline{9.7}&15.5&45.6&\textbf{5.4}&2.9&\underline{4.5}&13.1&12.6&11.8&\textbf{9.1}&\underline{4.3}&12.0&12.2\\\cline{2-16}

&${\mathcal{L}_{d_{i,j}}}{(\rm{{\textbf{s},{\textbf{t}}}})}$&ResNet50&9.8&13.2&26.7&6.5&\textbf{2.5}&\textbf{4.2}&\underline{9.4}&10.6&\textbf{11.0}&10.5&\textbf{4.2}&\textbf{9.8}&9.55\\

&$\mathcal{L}_{{\rm\textbf{D}}_{i,j}^2}{(\rm{{\textbf{s},{\textbf{t}}}})}$&ResNet50&10.5&12.6&23.1&5.8&\underline{2.6}&5.1&9.6&11.2&\underline{11.5}&9.7&\underline{4.3}&\underline{10.4}&9.47\\

&$\mathcal{L}_{{\rm\textbf{D}}_{i,j}^{H\tau}}{(\rm{{\textbf{s},{\textbf{t}}}})}$&ResNet50&11.3&\textbf{11.8}&\underline{19.2}&6.8&3.1&5.0&10.1&\textbf{9.8}&11.8&\underline{9.4}&4.7&11.2&\underline{9.46}\\

&$\mathcal{L}_{{\rm\textbf{D}}_{i,j}^2}{(\rm{{\textbf{s},\overline{\textbf{t}}}})}$&ResNet50&10.1&\underline{12.0}&21.4&\underline{5.6}&2.8&4.6&10.0&\underline{10.3}&12.3&9.6&4.5&11.6&\textbf{9.37}\\\hline

\end{tabular}\label{tab:5}
\end{center}
\caption{Results on PASCAL 3D+ view point estimation $w.r.t.$ $Acc_{\frac{\pi}{6}}$ $Acc_{\frac{\pi}{18}}$ (the higher the better) and $MedErr$ (the lower the better). Our results are based on ResNet50 backbone and without using external training data.}
\end{table*}

%% file: 5_Experiments3.tex
\subsection{3D object pose}

PASCAL3D+ dataset \cite{xiang2014beyond} consists of 12 common categorical rigid object images from the Pascal VOC12 and ImageNet dataset \cite{deng2009imagenet} with both detection and pose annotations. On average, about 3,000 object instances per category are captured in the wild, making it challenging for estimating object pose. We follow the typical experimental protocol, that using ground truth detection for both training and testing, and choosing Pascal validation set to evaluate our viewpoint estimation quality \cite{mahendran2018mixed,prokudin2018deep,grabner20183d}.

The pose of an object in 3D space is usually defined as a 3-tuple (azimuth, elevation, cyclo-rotation), and each of them is computed separately. We note that the range of elevation is [0,$\pi$], the Wasserstein distance for non-periodic ordered data can be computed via Eq. \eqref{con:ordinal}. We choose the Binomial-uniform mixture distribution ($\xi=0.2,\eta=0.05,K=20,p=0.5$) to construct our conservative label. The same data augmentation and ResNet50 from \cite{mahendran2018mixed} (mixture of CE and regression loss) is adopted for fair comparisons, which has 12 branches for each categorical and each branch has three softmax units for 3-tuple.

We consider two metrics commonly applied in literature: Accuracy at $\frac{\pi}{6}$, and Median error ($i.e.,$ the median of rotation angle error). Table \textcolor{red}{5} compares our approach with previous techniques. Our methods outperform previous approaches in both testing metrics. The improvements are more exciting than recent works. Specifically, $\small \mathcal{L}_{{\rm\textbf{D}}_{i,j}^2}{(\rm{{\textbf{s},\overline{\textbf{t}}}})}$ outperforms \cite{mahendran2018mixed} by 3.37\% in terms of $Acc_{\frac{\pi}{6}}$, and the reduces $MedErr$ from 10.4 \cite{zhou2018starmap} to 9.37 (by 9.5\%).  

Besides, we further evaluate $Acc\frac{\pi}{18}$, which assesses the percentage of more accurate predictions. This shows the prediction probabilities are closely distributed around the ground of truth pose.

%% file: 6_Conclusions.tex
\section{Conclusions}
We have introduced a simple yet efficient loss function for pose estimation, based on the Wasserstein distance. Its ground metric represents the class correlation and can be predefined using an increasing function of arc length or learned by alternative optimization. Both the outlier and inlier noise in pose data are incorporated in a unimodal-uniform mixture distribution to construct the conservative label. We systematically discussed the fast closed-form solutions in one-hot and conservative label cases. The results show that the best performance can be achieved by choosing convex function, Binomial distribution for smoothing and solving its exact solution. Although it was originally developed for pose estimation, it is essentially applicable to other problems with discrete and periodic labels. In the future, we plan to develop a more stable adaptive ground metric learning scheme for more classes, and adjust the shape of conservative target distribution automatically.

\section{Acknowledgement}
The funding support from Youth Innovation Promotion Association, CAS (2017264), Innovative Foundation of
CIOMP, CAS (Y586320150) and Hong Kong Government General Research Fund GRF (Ref. No.152202/14E) are greatly appreciated.

%% file: main.bbl
\begin{thebibliography}{10}\itemsep=-1pt

\bibitem{andriluka2010monocular}
Mykhaylo Andriluka, Stefan Roth, and Bernt Schiele.
\newblock Monocular 3d pose estimation and tracking by detection.
\newblock In {\em Computer Vision and Pattern Recognition (CVPR), 2010 IEEE
  Conference on}, pages 623--630. IEEE, 2010.

\bibitem{arjovsky2017wasserstein}
Martin Arjovsky, Soumith Chintala, and L{\'e}on Bottou.
\newblock Wasserstein gan.
\newblock {\em arXiv preprint arXiv:1701.07875}, 2017.

\bibitem{belagiannis2015robust}
Vasileios Belagiannis, Christian Rupprecht, Gustavo Carneiro, and Nassir Navab.
\newblock Robust optimization for deep regression.
\newblock In {\em Proceedings of the IEEE International Conference on Computer
  Vision}, pages 2830--2838, 2015.

\bibitem{beyer2015biternion}
Lucas Beyer, Alexander Hermans, and Bastian Leibe.
\newblock Biternion nets: Continuous head pose regression from discrete
  training labels.
\newblock In {\em German Conference on Pattern Recognition}, pages 157--168.
  Springer, 2015.

\bibitem{burkard2009society}
R Burkard, M Dell’Amico, and S Martello.
\newblock Society for industrial and applied mathematics, assignment problems.
  philadelphia (pa.): Siam.
\newblock {\em Society for Industrial and Applied Mathematics}, 2009.

\bibitem{cabrelli1995kantorovich}
Carlos~A Cabrelli and Ursula~M Molter.
\newblock The kantorovich metric for probability measures on the circle.
\newblock {\em Journal of Computational and Applied Mathematics},
  57(3):345--361, 1995.

\bibitem{cabrelli1998linear}
Carlos~A Cabrelli and Ursula~M Molter.
\newblock A linear time algorithm for a matching problem on the circle.
\newblock {\em Information processing letters}, 66(3):161--164, 1998.

\bibitem{cha2002fast}
Sung-Hyuk Cha.
\newblock A fast hue-based colour image indexing algorithm.
\newblock {\em Machine Graphics \& Vision International Journal},
  11(2/3):285--295, 2002.

\bibitem{cha2002measuring}
Sung-Hyuk Cha and Sargur~N Srihari.
\newblock On measuring the distance between histograms.
\newblock {\em Pattern Recognition}, 35(6):1355--1370, 2002.

\bibitem{che2019deep}
Tong Che, Xiaofeng Liu, Site Li, Yubin Ge, Ruixiang Zhang, Caiming Xiong, and
  Yoshua Bengio.
\newblock Deep verifier networks: Verification of deep discriminative models
  with deep generative models.
\newblock In {\em ArXiv}, 2019.

\bibitem{cuturi2013sinkhorn}
Marco Cuturi.
\newblock Sinkhorn distances: Lightspeed computation of optimal transport.
\newblock In {\em Advances in neural information processing systems}, pages
  2292--2300, 2013.

\bibitem{delon2010fast}
Julie Delon, Julien Salomon, and Andrei Sobolevski.
\newblock Fast transport optimization for monge costs on the circle.
\newblock {\em SIAM Journal on Applied Mathematics}, 70(7):2239--2258, 2010.

\bibitem{deng2009imagenet}
Jia Deng, Wei Dong, Richard Socher, Li-Jia Li, Kai Li, and Li Fei-Fei.
\newblock Imagenet: A large-scale hierarchical image database.
\newblock In {\em Computer Vision and Pattern Recognition, 2009. CVPR 2009.
  IEEE Conference on}, pages 248--255. Ieee, 2009.

\bibitem{elhoseiny2016comparative}
Mohamed Elhoseiny, Tarek El-Gaaly, Amr Bakry, and Ahmed Elgammal.
\newblock A comparative analysis and study of multiview cnn models for joint
  object categorization and pose estimation.
\newblock In {\em International Conference on Machine learning}, pages
  888--897, 2016.

\bibitem{fisher2005caviar}
Robert Fisher, Jose Santos-Victor, and James Crowley.
\newblock Caviar: Context aware vision using image-based active recognition,
  2005.

\bibitem{frogner2015learning}
Charlie Frogner, Chiyuan Zhang, Hossein Mobahi, Mauricio Araya, and Tomaso~A
  Poggio.
\newblock Learning with a wasserstein loss.
\newblock In {\em Advances in Neural Information Processing Systems}, pages
  2053--2061, 2015.

\bibitem{grabner20183d}
Alexander Grabner, Peter~M Roth, and Vincent Lepetit.
\newblock 3d pose estimation and 3d model retrieval for objects in the wild.
\newblock In {\em Proceedings of the IEEE Conference on Computer Vision and
  Pattern Recognition}, pages 3022--3031, 2018.

\bibitem{han2019unsupervised}
Ligong Han, Yang Zou, Ruijiang Gao, Lezi Wang, and Dimitris Metaxas.
\newblock Unsupervised domain adaptation via calibrating uncertainties.
\newblock In {\em Proceedings of the IEEE Conference on Computer Vision and
  Pattern Recognition Workshops}, pages 99--102, 2019.

\bibitem{hara2017growing}
Kota Hara and Rama Chellappa.
\newblock Growing regression tree forests by classification for continuous
  object pose estimation.
\newblock {\em International Journal of Computer Vision}, 122(2):292--312,
  2017.

\bibitem{hara2017designing}
Kota Hara, Raviteja Vemulapalli, and Rama Chellappa.
\newblock Designing deep convolutional neural networks for continuous object
  orientation estimation.
\newblock {\em arXiv preprint arXiv:1702.01499}, 2017.

\bibitem{he2016deep}
Kaiming He, Xiangyu Zhang, Shaoqing Ren, and Jian Sun.
\newblock Deep residual learning for image recognition.
\newblock In {\em Proceedings of the IEEE conference on computer vision and
  pattern recognition}, pages 770--778, 2016.

\bibitem{huang2017soft}
Dong Huang, Longfei Han, and Fernando De~la Torre.
\newblock Soft-margin mixture of regressions.
\newblock In {\em Proc. CVPR}, 2017.

\bibitem{huber2011robust}
Peter~J Huber.
\newblock Robust statistics.
\newblock In {\em International Encyclopedia of Statistical Science}, pages
  1248--1251. Springer, 2011.

\bibitem{kolouri2016sliced}
Soheil Kolouri, Yang Zou, and Gustavo~K Rohde.
\newblock Sliced wasserstein kernels for probability distributions.
\newblock In {\em Proceedings of the IEEE Conference on Computer Vision and
  Pattern Recognition}, pages 5258--5267, 2016.

\bibitem{liu2019research}
Xiaofeng Liu.
\newblock Research on the technology of deep learning based face image
  recognition.
\newblock In {\em Thesis}, 2019.

\bibitem{liu2018dependency}
Xiaofeng Liu, Kumar B.V.K, Chao Yang, Qingming Tang, and Jane You.
\newblock Dependency-aware attention control for unconstrained face recognition
  with image sets.
\newblock In {\em European Conference on Computer Vision}, 2018.

\bibitem{liu2019unimodalb}
Xiaofeng Liu, Fangfang Fan, Lingsheng Kong, Ping Jia, You Jane, and Jun Lu.
\newblock Unimodal regularized neuron stick-breaking for ordinal
  classification.
\newblock In {\em ArXiv}, 2019.

\bibitem{liu2018adaptive}
Xiaofeng Liu, Yubin Ge, Chao Yang, and Ping Jia.
\newblock Adaptive metric learning with deep neural networks for video-based
  facial expression recognition.
\newblock {\em Journal of Electronic Imaging}, 27(1):013022, 2018.

\bibitem{liu2019dependency}
Xiaofeng Liu, Zhenhua Guo, Jane Jia, and B.V.K. Kumar.
\newblock Dependency-aware attention control for image set-based face
  recognition.
\newblock In {\em IEEE TIFS}, 2019.

\bibitem{liu2019permutation}
Xiaofeng Liu, Zhenhua Guo, Site Li, Jane You, and Kumar B.V.K.
\newblock Permutation-invariant feature restructuring for correlation-aware
  image set-based recognition.
\newblock In {\em ICCV}, 2019.

\bibitem{liu2019unimodala}
Xiaofeng Liu, Xu Han, Yukai Qiao, Yi Ge, Site Li, and Jun Lu.
\newblock Unimodal-uniform constrained wasserstein training for medical
  diagnosis.
\newblock In {\em ICCV}, 2019.

\bibitem{liu2019hard}
Xiaofeng Liu, BVK~Vijaya Kumar, Ping Jia, and Jane You.
\newblock Hard negative generation for identity-disentangled facial expression
  recognition.
\newblock {\em Pattern Recognition}, 88:1--12, 2019.

\bibitem{liu2019feature}
Xiaofeng Liu, Site Li, Lingsheng Kong, Wanqing Xie, Ping Jia, Jane You, and BVK
  Kumar.
\newblock Feature-level frankenstein: Eliminating variations for discriminative
  recognition.
\newblock In {\em Proceedings of the IEEE Conference on Computer Vision and
  Pattern Recognition}, pages 637--646, 2019.

\bibitem{liu2018joint}
Xiaofeng Liu, Zhaofeng Li, Lingsheng Kong, Zhihui Diao, Junliang Yan, Yang Zou,
  Chao Yang, Ping Jia, and Jane You.
\newblock A joint optimization framework of low-dimensional projection and
  collaborative representation for discriminative classification.
\newblock In {\em 2018 24th International Conference on Pattern Recognition
  (ICPR)}, pages 1493--1498.

\bibitem{liu2017adaptive}
Xiaofeng Liu, BVK Vijaya~Kumar, Jane You, and Ping Jia.
\newblock Adaptive deep metric learning for identity-aware facial expression
  recognition.
\newblock In {\em CVPR Workshops}, pages 20--29, 2017.

\bibitem{liu2018ordinal}
Xiaofeng Liu, Yang Zou, Yuhang Song, Chao Yang, Jane You, and BV
  K~Vijaya~Kumar.
\newblock Ordinal regression with neuron stick-breaking for medical diagnosis.
\newblock In {\em Proceedings of the European Conference on Computer Vision
  (ECCV)}, pages 0--0, 2018.

\bibitem{mahendran2018mixed}
Siddharth Mahendran, Haider Ali, and Rene Vidal.
\newblock A mixed classification-regression framework for 3d pose estimation
  from 2d images.
\newblock {\em BMVC}, 2018.

\bibitem{massa2016crafting}
Francisco Massa, Renaud Marlet, and Mathieu Aubry.
\newblock Crafting a multi-task cnn for viewpoint estimation.
\newblock {\em British Machine Vision Conference}, 2016.

\bibitem{mousavian20173d}
Arsalan Mousavian, Dragomir Anguelov, John Flynn, and Jana Ko{\v{s}}eck{\'a}.
\newblock 3d bounding box estimation using deep learning and geometry.
\newblock In {\em Computer Vision and Pattern Recognition (CVPR), 2017 IEEE
  Conference on}, pages 5632--5640. IEEE, 2017.

\bibitem{murphy2009head}
Erik Murphy-Chutorian and Mohan~Manubhai Trivedi.
\newblock Head pose estimation in computer vision: A survey.
\newblock {\em IEEE transactions on pattern analysis and machine intelligence},
  31(4):607--626, 2009.

\bibitem{orlin1993faster}
James~B Orlin.
\newblock A faster strongly polynomial minimum cost flow algorithm.
\newblock {\em Operations research}, 41(2):338--350, 1993.

\bibitem{ozuysal2009pose}
Mustafa Ozuysal, Vincent Lepetit, and Pascal Fua.
\newblock Pose estimation for category specific multiview object localization.
\newblock In {\em Computer Vision and Pattern Recognition, 2009. CVPR 2009.
  IEEE Conference on}, pages 778--785. IEEE, 2009.

\bibitem{pavlakos20176}
Georgios Pavlakos, Xiaowei Zhou, Aaron Chan, Konstantinos~G Derpanis, and
  Kostas Daniilidis.
\newblock 6-dof object pose from semantic keypoints.
\newblock In {\em Robotics and Automation (ICRA), 2017 IEEE International
  Conference on}, pages 2011--2018. IEEE, 2017.

\bibitem{pele2008linear}
Ofir Pele and Michael Werman.
\newblock A linear time histogram metric for improved sift matching.
\newblock In {\em European conference on computer vision}, pages 495--508.
  Springer, 2008.

\bibitem{pereyra2017regularizing}
Gabriel Pereyra, George Tucker, Jan Chorowski, {\L}ukasz Kaiser, and Geoffrey
  Hinton.
\newblock Regularizing neural networks by penalizing confident output
  distributions.
\newblock {\em arXiv preprint arXiv:1701.06548}, 2017.

\bibitem{prokudin2018deep}
Sergey Prokudin, Peter Gehler, and Sebastian Nowozin.
\newblock Deep directional statistics: Pose estimation with uncertainty
  quantification.
\newblock {\em ECCV}, 2018.

\bibitem{rabin2009statistical}
Julien Rabin, Julie Delon, and Yann Gousseau.
\newblock A statistical approach to the matching of local features.
\newblock {\em SIAM Journal on Imaging Sciences}, 2(3):931--958, 2009.

\bibitem{rad2017bb8}
Mahdi Rad and Vincent Lepetit.
\newblock Bb8: A scalable, accurate, robust to partial occlusion method for
  predicting the 3d poses of challenging objects without using depth.
\newblock In {\em International Conference on Computer Vision}, volume~1,
  page~5, 2017.

\bibitem{raza2018appearance}
Mudassar Raza, Zonghai Chen, Saeed-Ur Rehman, Peng Wang, and Peng Bao.
\newblock Appearance based pedestrians’ head pose and body orientation
  estimation using deep learning.
\newblock {\em Neurocomputing}, 272:647--659, 2018.

\bibitem{rizzo2016energy}
Maria~L Rizzo and G{\'a}bor~J Sz{\'e}kely.
\newblock Energy distance.
\newblock {\em Wiley Interdisciplinary Reviews: Computational Statistics},
  8(1):27--38, 2016.

\bibitem{rubner2000earth}
Yossi Rubner, Carlo Tomasi, and Leonidas~J Guibas.
\newblock The earth mover's distance as a metric for image retrieval.
\newblock {\em International journal of computer vision}, 40(2):99--121, 2000.

\bibitem{ruschendorf1985wasserstein}
Ludger R{\"u}schendorf.
\newblock The wasserstein distance and approximation theorems.
\newblock {\em Probability Theory and Related Fields}, 70(1):117--129, 1985.

\bibitem{simonyan2014very}
Karen Simonyan and Andrew Zisserman.
\newblock Very deep convolutional networks for large-scale image recognition.
\newblock {\em arXiv preprint arXiv:1409.1556}, 2014.

\bibitem{su2017order}
Bing Su and Gang Hua.
\newblock Order-preserving wasserstein distance for sequence matching.
\newblock In {\em IEEE Conference on Computer Vision and Pattern Recognition},
  pages 2906--2914, 2017.

\bibitem{su2015render}
Hao Su, Charles~R Qi, Yangyan Li, and Leonidas~J Guibas.
\newblock Render for cnn: Viewpoint estimation in images using cnns trained
  with rendered 3d model views.
\newblock In {\em Proceedings of the IEEE International Conference on Computer
  Vision}, pages 2686--2694, 2015.

\bibitem{szegedy2016rethinking}
Christian Szegedy, Vincent Vanhoucke, Sergey Ioffe, Jon Shlens, and Zbigniew
  Wojna.
\newblock Rethinking the inception architecture for computer vision.
\newblock In {\em Proceedings of the IEEE conference on computer vision and
  pattern recognition}, pages 2818--2826, 2016.

\bibitem{szeto2017click}
Ryan Szeto and Jason~J Corso.
\newblock Click here: Human-localized keypoints as guidance.
\newblock {\em arXiv preprint arXiv:1703.09859}, 2017.

\bibitem{tulsiani2015viewpoints}
Shubham Tulsiani and Jitendra Malik.
\newblock Viewpoints and keypoints.
\newblock In {\em Proceedings of the IEEE Conference on Computer Vision and
  Pattern Recognition}, pages 1510--1519, 2015.

\bibitem{villani2003topics}
C{\'e}dric Villani.
\newblock {\em Topics in optimal transportation}.
\newblock Number~58. American Mathematical Soc., 2003.

\bibitem{wang2016viewpoint}
Yumeng Wang, Shuyang Li, Mengyao Jia, and Wei Liang.
\newblock Viewpoint estimation for objects with convolutional neural network
  trained on synthetic images.
\newblock In {\em Pacific Rim Conference on Multimedia}, pages 169--179.
  Springer, 2016.

\bibitem{werman1986bipartite}
Michael Werman, Shmuel Peleg, Robert Melter, and T~Yung Kong.
\newblock Bipartite graph matching for points on a line or a circle.
\newblock {\em Journal of Algorithms}, 7(2):277--284, 1986.

\bibitem{wu2016single}
Jiajun Wu, Tianfan Xue, Joseph~J Lim, Yuandong Tian, Joshua~B Tenenbaum,
  Antonio Torralba, and William~T Freeman.
\newblock Single image 3d interpreter network.
\newblock In {\em European Conference on Computer Vision}, pages 365--382.
  Springer, 2016.

\bibitem{xiang2014beyond}
Yu Xiang, Roozbeh Mottaghi, and Silvio Savarese.
\newblock Beyond pascal: A benchmark for 3d object detection in the wild.
\newblock In {\em Applications of Computer Vision (WACV), 2014 IEEE Winter
  Conference on}, pages 75--82. IEEE, 2014.

\bibitem{yang2018hierarchical}
Dan Yang, Yanlin Qian, Ke Chen, Eleni Berki, and Joni-Kristian
  K{\"a}m{\"a}r{\"a}inen.
\newblock Hierarchical sliding slice regression for vehicle viewing angle
  estimation.
\newblock {\em IEEE Transactions on Intelligent Transportation Systems},
  19(6):2035--2042, 2018.

\bibitem{zhou2018starmap}
Xingyi Zhou, Arjun Karpur, Linjie Luo, and Qixing Huang.
\newblock Starmap for category-agnostic keypoint and viewpoint estimation.
\newblock {\em ECCV}, 2018.

\bibitem{zou2019confidence}
Yang Zou, Zhiding Yu, Xiaofeng Liu, Jingsong Wang, and Kumar B.V.K.
\newblock Confidence regularized self-training.
\newblock {\em ICCV}, 2019.

\end{thebibliography}
